\pdfoutput=1

\documentclass[11pt]{article}

\usepackage[final]{acl}

\usepackage{times}
\usepackage{latexsym}

\usepackage[T1]{fontenc}

\usepackage[utf8]{inputenc}

\usepackage{microtype}
\usepackage{algorithm}
\usepackage{subcaption}
\usepackage{multirow}
\usepackage{multicol}
\usepackage{algpseudocode}
\usepackage{amssymb}
\usepackage{longtable, booktabs}
\usepackage{hyperref}
\usepackage{colortbl}

\usepackage{inconsolata}
\usepackage{booktabs}

\usepackage{graphicx}

\usepackage{mathtools}

\usepackage{amsmath,amsfonts,bm}

\def\eqref#1{equation~\ref{#1}}

\def\1{\bm{1}}

\def\rvx{{\mathbf{x}}}
\def\rvy{{\mathbf{y}}}

\def\ervy{{\textnormal{y}}}

\DeclareMathAlphabet{\mathsfit}{\encodingdefault}{\sfdefault}{m}{sl}
\SetMathAlphabet{\mathsfit}{bold}{\encodingdefault}{\sfdefault}{bx}{n}

\DeclareMathOperator*{\argmax}{arg\,max}

\title{Inference-Time Language Model Alignment via Integrated Value Guidance}

\author{
  Zhixuan Liu\textsuperscript{1,2}\thanks{Contribute equally.}~, Zhanhui Zhou\textsuperscript{1}\footnotemark[1]~, Yuanfu Wang\textsuperscript{1}, Chao Yang\textsuperscript{1}\thanks{Corresponding author.}~, 
  Yu Qiao\textsuperscript{1} \\
  \textsuperscript{1}Shanghai Artificial Intelligence Laboratory \\
  \textsuperscript{2}Shanghai Jiaotong University \\
  \texttt{\{liuzhixuan, zhouzhanhui, wangyuanfu, yangchao, qiaoyu\}@pjlab.org.cn}
}

\begin{document}

\maketitle
\begin{abstract}

Large language models are typically fine-tuned to align with human preferences, but tuning large models is computationally intensive and complex. 
In this work, we introduce \textit{Integrated Value Guidance} (IVG), a method that uses implicit and explicit value functions to guide language model decoding at token and chunk-level respectively, efficiently aligning large language models purely at inference time.
This approach circumvents the complexities of direct fine-tuning and outperforms traditional methods.
Empirically, we demonstrate the versatility of IVG across various tasks. In controlled sentiment generation and summarization tasks, our method significantly improves the alignment of large models using inference-time guidance from \texttt{gpt2}-based value functions. Moreover, in a more challenging instruction-following benchmark AlpacaEval 2.0, we show that both specifically tuned and off-the-shelf value functions greatly improve the length-controlled win rates of large models against \texttt{gpt-4-turbo} (e.g., $19.51\% \rightarrow 26.51\%$ for \texttt{Mistral-7B-Instruct-v0.2} and $25.58\% \rightarrow 33.75\%$ for \texttt{Mixtral-8x7B-Instruct-v0.1} with Tulu guidance).

\end{abstract}

\section{Introduction}

Learning-based algorithms have become the standard for aligning large language models (LLMs) with human preferences, as evidenced by numerous studies~\citep{ziegler2019fine, stiennon2020learning, ouyang2022training, rafailov2024direct, azar2024general}. Despite their success, fine-tuning LLMs is notably resource-intensive and poses implementation challenges~\citep{rafailov2024direct}. These challenges have catalyzed the development of inference-time alignment methods that maintain LLMs in a frozen state and guide their decoding during testing~\citep{mitchell2023emulator, liu2024tuning, mudgal2023controlled, kim2023critic, huang2024deal, gao2023scaling, beirami2024theoretical}. 

\begin{figure}[t]
    \centering
    \includegraphics[width=0.48\textwidth]{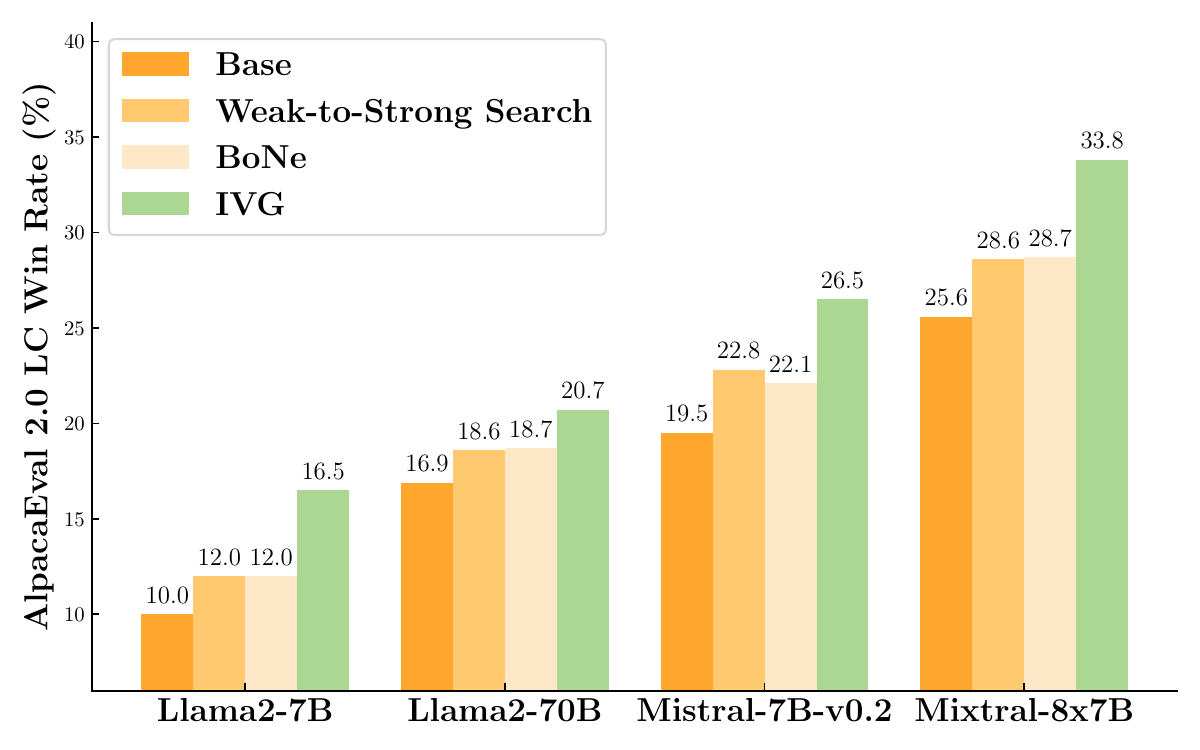}
    \caption{Illustration of Integrated Value Guidance (IVG) with parameters $W,K,L=2,2,30$. Weak-to-Strong Search~\citep{zhou2024weak} denotes the results with the same parameters. BoNe denotes Best-of-N Sampling ($N=4$) with explicit values. }
    \label{fig:irlmg_illustration}
    \vspace{-15pt}
\end{figure}

Value functions, which assess the quality or alignment of generated text with desired criteria, have proven effective for inference-time alignment in two primary forms: (1) implicit value functions, represented by the log-probability differences between fine-tuned and base models~\citep{rafailov2024r, mitchell2023emulator, liu2024tuning, zhou2024emulated, liu2021dexperts}, and (2) explicit value functions, developed through direct training~\citep{mudgal2023controlled, yang-klein-2021-fudge}. 
Our empirical analysis reveals a significant performance discrepancy between these functions at different granularity levels of inference alignment (Section \ref{sec:exp}): explicit value functions excel at chunk-level evaluation, whereas implicit value functions are more effective at the token-level manipulation.

Recognizing this performance discrepancy, we introduce a novel algorithm called Integrated Value Guidance (IVG) to harness the strengths of both value function types. 
IVG combines the strengths of implicit and explicit value functions by applying implicit value functions to token-level sampling and explicit value functions to chunk-level beam search~\citep{mudgal2023controlled, zhou2024weak}.
IVG offers two significant advancements: (1) It integrates the distinctive performances of the two value function types across different granular strategies, thereby validating our theoretical models through empirical tests. (2) It introduces a robust inference-time alignment method, outperforming similar existing techniques in various evaluations. Figure \ref{fig:irlmg_illustration} illustrates the IVG method.

Empirically, IVG demonstrates its versatility in tasks such as controlled-sentiment generation~\citep{maas-EtAl:2011:ACL-HLT2011} and summarization~\citep{stiennon2020learning}, where using small models like \texttt{gpt2} with 124M parameters is effective in guiding larger models from the GPT-2 series~\citep{radford2019language} to achieve competitive results.
Further, in challenging instruction-following benchmarks such as \textit{AlpacaEval 2.0}~\citep{dubois2024length}, both open-source models (e.g., Tulu guidance) and our fully trained models (e.g., Ultra guidance) significantly enhance the length-controlled win rates of larger models against competitors like \texttt{gpt-4-turbo} (e.g., \texttt{Mistral-7B-Instruct-v0.2}~\citep{jiang2023mistral} from $19.51$ to $26.51$ and \texttt{Mixtral-8x7B\-Instruct-v0.1}~\citep{mixtral2023} from $25.58$ to $33.75$).

\section{Related Work}
Large unsupervised language models, trained on vast internet-scale datasets, have demonstrated remarkable capabilities~\citep{chowdhery2023palm, brown2020language, touvron2023llama}. Nonetheless, aligning these models with human values remains challenging. Traditionally, alignment is achieved through fine-tuning based on human evaluations of model-generated responses~\citep{ziegler2019fine, stiennon2020learning, ouyang2022training, rafailov2024direct, touvron2023llama2, llama3modelcard, bai2022training, bai2022constitutional}. While effective, this method demands significant computational and engineering resources. Moreover, the diversity of human values complicates the creation of universally aligned models~\citep{ouyang2022training, zhou2024beyond, mudgal2023controlled, rame2024rewarded, jang2023personalized, wang2024arithmetic}.

In response to these challenges, we propose an inference-time alignment approach that freezes pre-trained models while modulating their outputs through a decoding phase managed by smaller, specialized models. This strategy minimizes the need for extensive retraining and adapts more readily to individual preferences.

The conceptual framework for aligning language models at inference time is rooted in the use of value functions. Implicit value functions, as described by \citet{rafailov2024r}, proposed a token-level Markov Decision Process to adjust language model outputs based on the log-likelihood differences between fine-tuned and base models. \citet{mudgal2023controlled} introduced a method that leverages explicit value functions trained through a KL-regularized reinforcement learning objective, which acts as a prefix scorer. Our empirical findings, detailed in Section \ref{sec:exp}, underscore that while implicit value functions excel at refining token-level nuances, explicit value functions provide superior sequence-level contextual understanding.

These insights motivate our method, which integrates these value functions to enhance model alignment with human preferences during the decoding phase of model output generation.
\begin{figure*}[ht!]
    \centering
    \includegraphics[width=1.0\textwidth, trim={3cm 4cm 3.5cm 4cm},clip,scale=2.0]{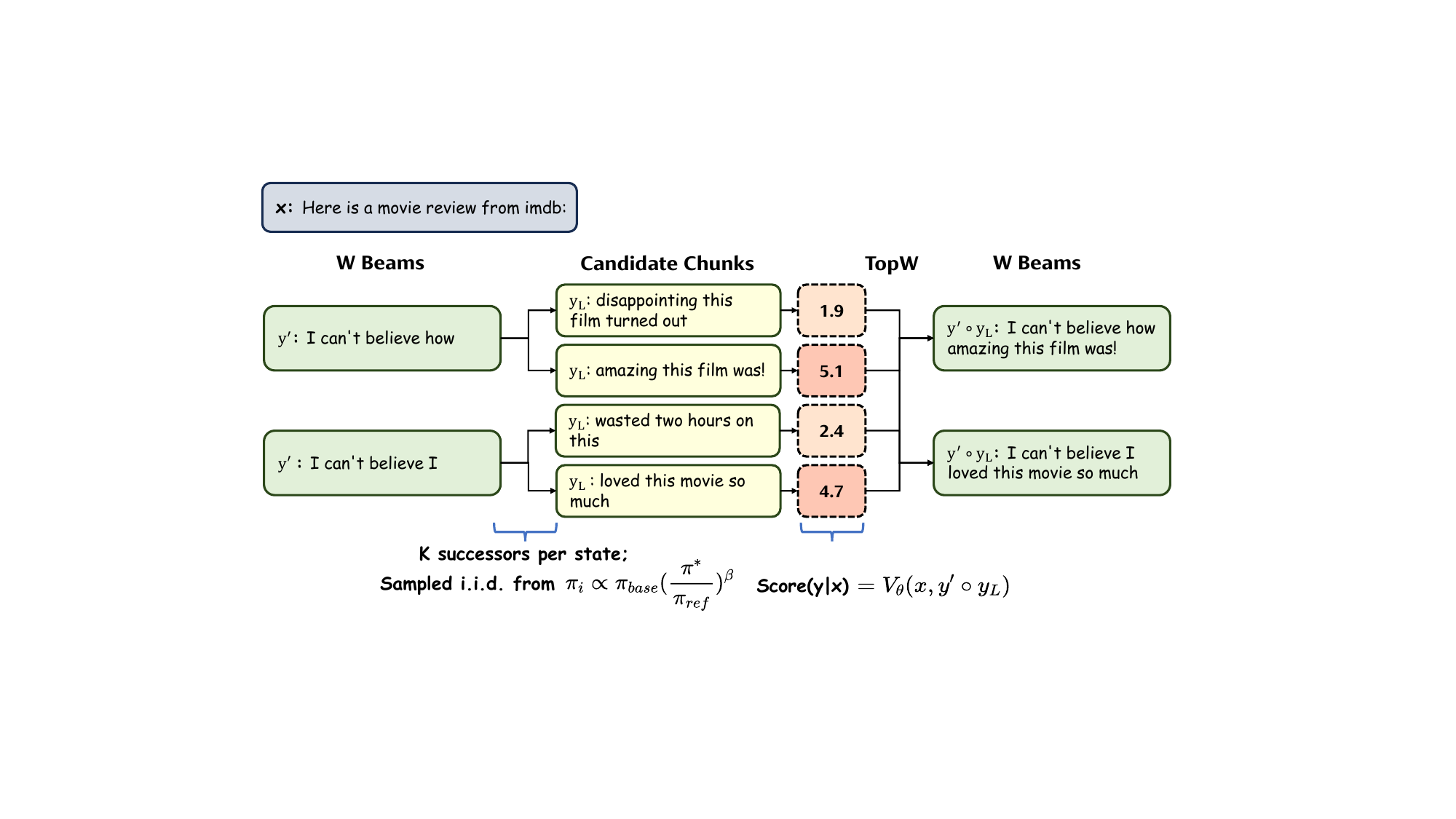}
    \caption{Illustration of Integrated Value Guidance (IVG) with beam width $W=2$, successors per state $K=2$, and chunk length $L=5$. }

    \label{fig:irlmg_illustration}
\end{figure*}

\section{Preliminaries}\label{sec:preliminaries}
In this section, we introduce the mathematical formulation of aligning large language models (LLMs) with human preferences and then describe the implicit and explicit value functions used in our approach.

\subsection{Large Language Model Alignment with Human Preferences}

Aligning large language models is commonly formulated as a Kullback-Leibler (KL)-constrained optimization problem~\citep{ziegler2019fine}:
\begin{equation}
\label{eq:obj}
\begin{aligned}
    \argmax_{\pi} \, & \mathbb{E}_{\rvx \sim p(\rvx), \rvy \sim \pi(\rvy \mid \rvx)} \Big[ r(\rvx, \rvy) \\
    & \quad - \mathbb{D}_{\mathrm{KL}}\left(\pi(\rvy \mid \rvx) \| \pi_{\mathrm{ref}}(\rvy \mid \rvx)\right) \Big],
\end{aligned}
\end{equation}
where \( p(\rvx) \) denotes the distribution of prompts, \( \rvy \) denotes the responses generated by the language model, \( r \) is the preference reward function induced from a preference datasets \( \mathcal{D} = \{(\rvx, \rvy_w, \rvy_l) \} \), and \( \mathbb{D}_{\mathrm{KL}} \) denotes the KL-divergence that constrains deviations from a reference model \( \pi_{\mathrm{ref}} \). 

\subsection{Implicit and Explicit Value Functions}

The value function estimates the expected terminal reward \( r(\rvx, \rvy) \) when following the policy \( \pi \) from a given state \( (\rvx, \rvy_{\leq t}) \):
\begin{equation}
\label{eq:value}
V(\rvx, \rvy_{\leq t}) = \mathbb{E}_{\rvy \sim \pi(\cdot \mid \rvx, \rvy_{\leq t})} \left[ r(\rvx, \rvy) \right].
\end{equation}

Different forms of the value function and optimization objectives lead to various interpretations. In this work, we demonstrate both the implicit and explicit forms:

\paragraph{Implicit Value Function.}
    The implicit value function is defined under the optimal policy \( \pi^* \) and is derived from the differences in log probabilities between the tuned model \( \pi^* \) and the reference model \( \pi_{\text{ref}} \), as achieved by an alignment algorithm such as Direct Preference Optimization (DPO)~\citep{rafailov2024direct}.    
    Specifically, the implicit value function evaluating a partial response sequence \( \rvy_{\leq t} \) is given by:
    \begin{equation}
    V^*(\rvx, \rvy_{\leq t}) - V^*(\rvx) = 
    \log \frac{
    \pi^*(\rvy_{\leq t} \mid \rvx)
    }{
    \pi_{\text{ref}}(\rvy_{\leq t} \mid \rvx)
    }.
    \end{equation}

\paragraph{Explicit Value Function.}
The explicit value function is a directly trained prefix scorer, which can be approximated via maximum likelihood estimation on the offline preference dataset.
Specifically, the explicit value function is represented as:
    \begin{equation}
    V^*(\rvx, \rvy_{\leq t}) = V_{\theta}(\rvx, \rvy_{\leq t}),
    \end{equation}
    where \( V_{\theta} \) is the value function parameterized by \( \theta \).


\section{Method}\label{sec:method}

In this section, we introduce our proposed method, Integrated Value Guidance (IVG). First, we discuss how to utilize the value function in two distinct ways: token-wise sampling~\citep{mudgal2023controlled} and chunk-level beam search~\citep{zhou2024weak}. 
Next, we explain how to train the implicit and explicit value functions on the preference dataset. Finally, we present the overall inference-time alignment process and analyze the computational efficiency of our method.

\subsection{Value Function Guided Sampling and Search Strategies}

Given a value function, we can guide the sampling and search strategies in two ways: token-wise sampling and chunk-level beam search.

\subsubsection{Token-wise Sampling}

In the token-wise sampling strategy, we use the value function to adjust the sampling distribution of the next token. Specifically, we sample the next token \( \ervy_{t} \) according to the following distribution:
\begin{equation}
\label{eq:token-wise}
\begin{aligned}
    & \pi(\ervy_t \mid \rvx, \rvy_{\leq t-1})  \propto \\
    & \pi_{\text{base}}(\ervy_t \mid \rvx, \rvy_{\leq t-1}) \exp( \beta (V(\rvx, \rvy_{\leq t}) - \\
    & \qquad \qquad \qquad \qquad \qquad \beta (V(\rvx, \rvy_{\leq t-1})) ),
\end{aligned}
\end{equation}
where \( \pi_{\text{base}}(\ervy_t \mid \rvx, \rvy_{\leq t-1}) \) is the base distribution of the next token, and \( \beta \) is a hyperparameter that controls the strength of the value function guidance.

For the implicit value function, we have:
\begin{equation}
\label{eq:implicit-token-wise}
\begin{aligned}
    & \pi_i(\ervy_t \mid \rvx, \rvy_{\leq t-1}) \propto \\
    & \pi_{\text{base}}(\ervy_t \mid \rvx, \rvy_{\leq t-1}) \left( \frac{\pi^*(\ervy_t \mid \rvx, \rvy_{\leq t-1})}{\pi_{\text{ref}}(\ervy_t \mid \rvx, \rvy_{\leq t-1})} \right)^{\beta}.
\end{aligned}
\end{equation}

For the explicit value function, we have:
\begin{equation}
\label{eq:explicit-token-wise}
\begin{aligned}
    & \pi_e(\ervy_t \mid \rvx, \rvy_{\leq t-1}) \propto \\
    & \pi_{\text{base}}(\ervy_t \mid \rvx, \rvy_{\leq t-1}) \exp\left( \beta V_{\theta}(\rvx, \rvy_{\leq t}) \right).
\end{aligned}
\end{equation}

\subsubsection{Chunk-level Beam Search}

For search-based generation, the chunk-level beam search strategy is effective for value function-guided sampling. 
Previous research~\citep{zhou2024weak} has shown that chunk-level beam search outperforms best-of-N (BoN) sampling and requires an effective value function to rank candidate sequences. 
Specifically, we rank candidate sequences according to the following score:
\begin{equation}
\label{eq:beam-search}
\begin{aligned}
    r(\rvy_{\leq t}| \rvx) & \propto V^*(\rvx, \rvy_{\leq t}) - V^*(\rvx),
\end{aligned}
\end{equation}
where \( r(\rvy_{\leq t}| \rvx) \) is the score of the candidate sequence \( \rvy_{\leq t} \), and \( V^*(\rvx, \rvy_{\leq t}) \) and \( V^*(\rvx) \) are the expected value of the candidate sequence and the prefix, respectively.

For the implicit value function, we have:
\begin{equation}
\label{eq:implicit-beam-search}
\begin{aligned}
    r_i(\rvy_{\leq t}| \rvx) & = \log \frac{\pi^*(\rvy_{\leq t} | \rvx)}{\pi_{\text{ref}}(\rvy_{\leq t} | \rvx)}.
\end{aligned}
\end{equation}

For the explicit value function, we have:
\begin{equation}
\label{eq:explicit-beam-search}
\begin{aligned}
    r_e(\rvy_{\leq t}| \rvx) & = V_{\theta}(\rvx, \rvy_{\leq t}).
\end{aligned}
\end{equation}

\subsection{Training the Implicit and Explicit Value Functions}

There are various methods to train the implicit and explicit value functions on the preference dataset. 
For the implicit value function, we derive it from the difference in log probabilities between the tuned and untuned models, regardless of how the model was trained. 
For the explicit value function, we can employ any reinforcement learning algorithm. 

In this work, we use Direct Preference Optimization (DPO)~\citep{rafailov2024direct} to train the implicit value function and FUDGE~\citep{mudgal2023controlled, yang-klein-2021-fudge} to train the explicit value function.

For the implicit value function, we have~\citep{rafailov2024direct}:
\begin{equation}
\label{eq:implicit-training}
\begin{aligned}
    \ell_{\mathcal{F}}(\mathbf{x}, \mathbf{y}^{w}, \mathbf{y}^{l}) = - \log \sigma \Bigg( &\beta \log \frac{\pi_{\theta}(\mathbf{y}^{w} \mid \mathbf{x})}{\pi_{\text{ref}}(\mathbf{y}^{w} \mid \mathbf{x})} \\
    - &\beta \log \frac{\pi_{\theta}(\mathbf{y}^{l} \mid \mathbf{x})}{\pi_{\text{ref}}(\mathbf{y}^{l} \mid \mathbf{x})} \Bigg).
\end{aligned}
\end{equation}

For the explicit value function, we have~\citep{mudgal2023controlled}:
\begin{equation}
\label{eq:explicit-training}
\ell_{\mathcal{F}}(\mathbf{x}, \mathbf{y}; \theta) = \frac{1}{2} \sum_{t \in [|\mathbf{y}|]} \left( V_{\theta}(\mathbf{x}, \mathbf{y}_{\leq t}) - r(\mathbf{x}, \mathbf{y}) \right)^{2}.
\end{equation}




\subsection{Integrated Value Guidance}

The token-wise sampling and chunk-level beam search strategies can be combined to enhance the alignment of large language models with human preferences. Specifically, token-wise sampling adjusts the sampling distribution of the next token using the implicit value function, while chunk-level beam search ranks candidate sequences using the explicit value function. 

The IVG algorithm is illustrated in Figure~\ref{fig:irlmg_illustration}. The key insight is that by applying the implicit value function at the token level and the explicit value function at the chunk level, we effectively leverage the strengths of both. Compared to Weak-to-Strong Search~\citep{zhou2024weak}, we sample tokens from a policy adjusted by the implicit value function rather than the base policy and use the explicit value function to rank candidate sequences. Empirically, we find that this combination leads to better alignment with human preferences. We demonstrate the effectiveness of IVG in the following sections.

\subsection{Implementation and Complexity}

We analyze the implementation efficiency and computational complexity of the IVG method. At each time step, the main components contributing to the time complexity of IVG include:
(1) The base model performs a forward pass to compute the probability distribution of the next token based on the given context.
(2) The implicit value functions (including $\pi^*$ and $\pi_{\text{ref}}$) perform forward passes to compute the probability distributions for the next token and calculate their difference. This difference is then combined with the base model's probability distribution to obtain the final next token distribution.
(3) When the current chunk reaches the length $L$, we compute the value of all candidate sequences using the explicit value function $V_\theta$, and select the top-$W$ sequences.



  

Consider the complexity of a single decoding step with a context of length $t$ tokens. The complexity is:
\begin{itemize}
    \item If the current chunk length $\neq L$:
    \[
    T(t) = T_{\text{base}}(t) + T_{\pi^*}(t) + T_{\pi_{\text{ref}}}(t).
    \]
    \item If the current chunk length = $L$:
    \[
    T(t) =  T_{\text{base}}(t) + T_{\pi^*}(t) + T_{\pi_{\text{ref}}}(t) + T_{V_\theta}(t). 
    \]
\end{itemize}


To simplify, consider the case where $L=1$. The total time complexity becomes:
\[
T(t) =  T_{\text{base}}(t) + T_{\pi^*}(t) + T_{\pi_{\text{ref}}}(t) + T_{V_\theta}(t).
\]
Here, $T_{\text{base}}(t)$ and other terms represent the inference time of the respective models for a sequence of length $t$. Initially, due to pairwise attention computations, $T(t) = O(t^2)$. However, during generation, we can cache previous computations, reducing the complexity to $O(t)$.
Therefore, the per-step inference complexity is:
\[
T(t) = O(t) \times (C_{\text{base}} + C_{\pi^*} + C_{\pi_{\text{ref}}} + C_{V_\theta}),
\]
where $C_{\text{base}}$, $C_{\pi^*}$, $C_{\pi_{\text{ref}}}$, and $C_{V_\theta}$ are constants representing the computational costs of each model.

In summary, while IVG does not change the asymptotic time complexity of the generation process, it introduces additional computational overhead due to multiple forward passes and increased memory usage. We will empirically evaluate the computational complexity of different methods in the experimental section.





\section{Experiments}\label{sec:exp}

In this section, we empirically evaluate the ability of the proposed IVG to align large language models with human preferences using only inference-time guidance from small language models. 
First, in \textbf{controlled-sentiment generation}~\citep{maas-EtAl:2011:ACL-HLT2011} and \textbf{summarization}~\citep{stiennon2020learning},
we tune \texttt{gpt2} to model the desired behaviors in each task to get the implicit value function and train the explicit value function based on the same base model. Then, we use the trained implicit and explicit value functions to steer larger models of various scales (Section~\ref{subsec:exp-sentiment-summarization}). 

Next, in a more difficult \textbf{instruction-following} benchmark, AlpacaEval 2.0~\citep{dubois2024length}, in addition to tuning small models, we reuse the off-the-shelf open-source 7B models and their untuned versions as the implicit value function and 
train the explicit value function by one of the best-performance sequence-wise reward models evaluated by RewardBench~\citep{RewardBench}. We then use them to steer a series of large models.

\begin{figure*}[ht!]
    \centering
    \includegraphics[width=0.9\textwidth]{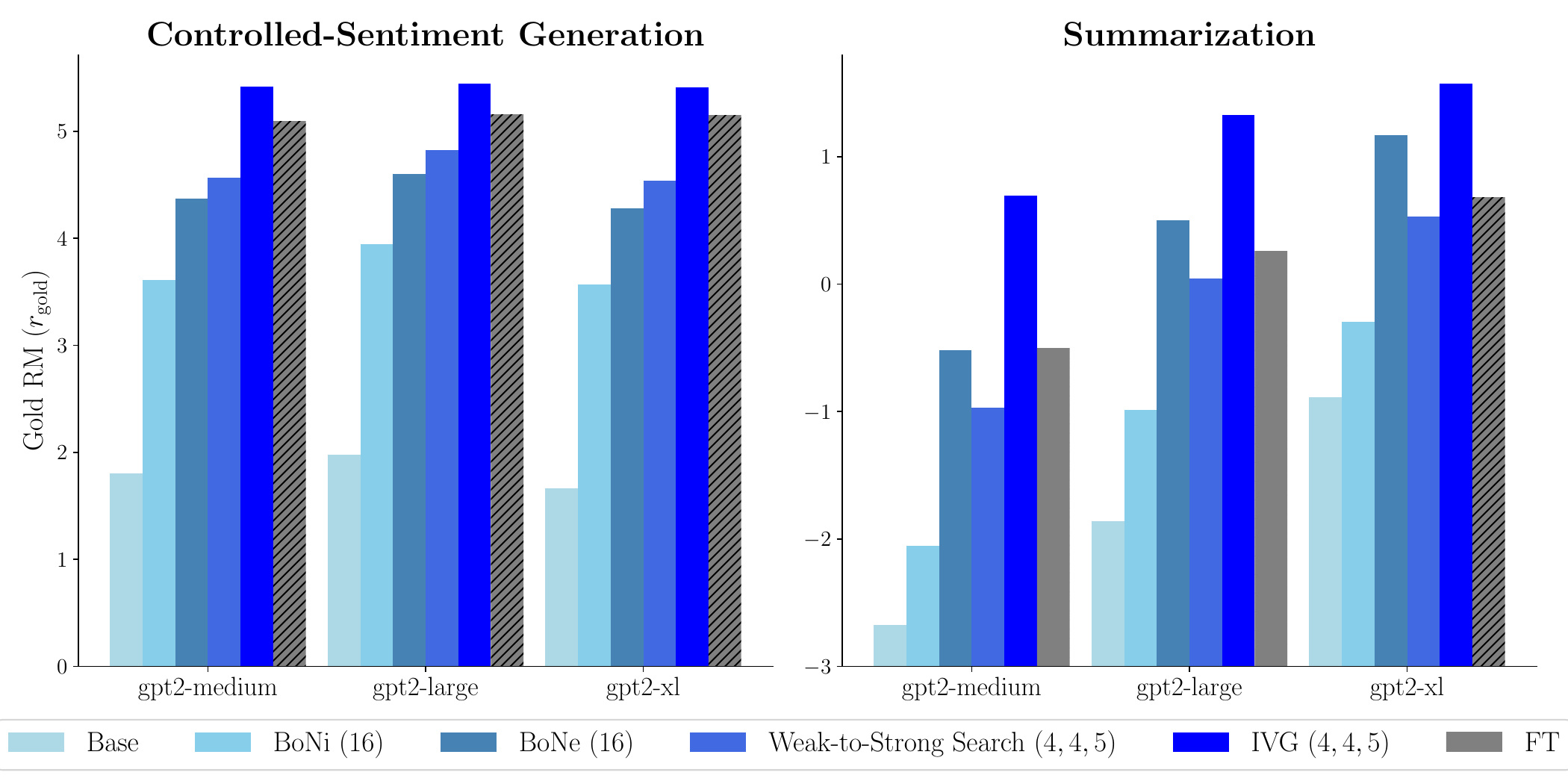}
    \caption{
        \textbf{The gold reward achieved for different large pre-trained models under the gpt2 guidance.} 
        We show the mean reward across three random seeds.
        BoNi and BoNe denote BoN ($N=16$) with implicit and explicit rewards, respectively;
        EFT ($\beta^*$) denotes the best EFT results among $\beta \in \{0.25, 0.5, 1, 2\} $; 
        CBS denotes the results with $W,K,L=4,4,5$ and implicit rewards; 
        IVG denotes the best results with $W,K,L=4,4,5$ among $\beta \in \{0.25, 0.5, 1, 2\} $.
    }
    \label{fig:performance_of_imdb_summarization}

\end{figure*}
\begin{figure*}[ht!]
    \centering
    \includegraphics[width=\textwidth]{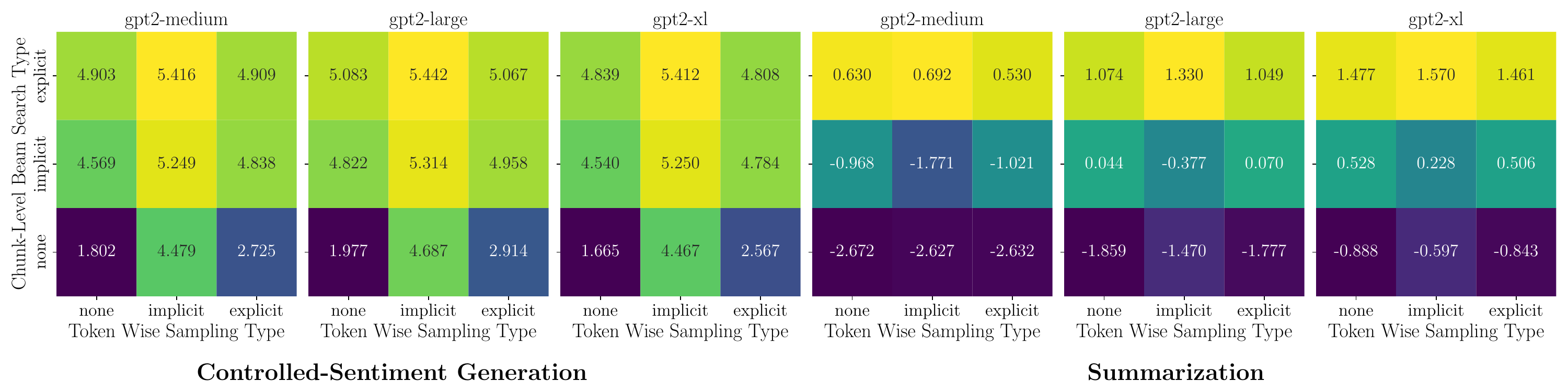}
    \caption{
        \textbf{The performance of different combinations of implicit and explicit value functions for token-wise sampling and chunk-level beam search in controlled-sentiment generation and summarization.}
        "implicit" and "explicit" denotes applying implicit and explicit value functions.
        "none" denotes the base model without any guidance.
        The number denotes the gold reward for the corresponding combination.
    }
    \label{fig:heatmap_of_imdb_summarization}

\end{figure*}

\paragraph{Baselines.}
Considering that some existing methods could be represented as special cases of the combinations of implicit and explicit value functions for token-wise sampling and chunk-level beam search, we compare the proposed IVG and different combinations of implicit and explicit value functions with the following baselines: (1) \textbf{Base}: the base model without any value function guidance; (2) \textbf{Best-of-N Sampling (BoN)}: BoNi uses $r = \log \pi^*(\rvy \mid \rvx) - \log \pi_{\text{ref}}(\rvy \mid \rvx)$ as rewards and BoNe uses $r = V_{\theta}(\rvx, \rvy)$ as rewards to select the highest-scoring responses among the $N$ independent response from the frozen base language model; (3) \textbf{FT}: fine-tuning the base model on the preference dataset.

Note that many existing methods can be represented by the framework: Emulator Fine-Tuning (EFT)~\citep{mitchell2023emulator} can be viewed as applying only the implicit value function in token-wise sampling. Weak-to-strong search can be viewed as applying only the implicit value function in chunk-level beam search. We use the EFTi, EFTe, CBSi and CBSe to represent the corresponding combination. For example, EFTi denotes applying implicit value function for token-wise sampling, and CBSe denotes applying explicit value function for chunk-level beam search.

\subsection {Controlled-Sentiment Generation \& Summarization}\label{subsec:exp-sentiment-summarization}
\paragraph{Setup.} For these two tasks, we follow the synthetic setups from \cite{gao2023scaling, lightman2023let, rafailov2024direct}, assuming \textbf{access to a gold reward model $r_{\text{gold}}$}. 
For controlled-sentiment generation, $r_{\text{gold}}$ encourages positive continuations of movie reviews, while for summarization, it encourages high-quality summaries of Reddit posts.
We \textbf{generate synthetic preference datasets} $\mathcal{D} = \{(\rvx, \rvy^w, \rvy^l)_i \}_{i=1}^N$ from $r_{\text{gold}}$ with $p(\rvy^1 \succ \rvy^2 \mid \rvx) = \sigma (r_{\text{gold}}(\rvx, \rvy^1) - r_{\text{gold}}(\rvx, \rvy^2))$ to mimic human feedback~\cite{bt}.

To \textbf{obtain the implicit value function}, we optimize \texttt{gpt2} (124M parameters) using the standard DPO pipeline~\citep{rafailov2024direct}: (1) we first obtain the reference model $\pi_{\text{ref}}$ through supervised fine-tuning on both chosen and rejected responses from the synthetic preference dataset, then (2) we apply DPO on the synthetic preference dataset with $\pi_{\text{ref}}$ as the reference policy to obtain the optimal language model $\pi^*$. 

To \textbf{obtain the explicit value function}, we train a prefix scorer $V_{\theta}$ using the FUDGE\citep{mudgal2023controlled,yang-klein-2021-fudge} algorithm on the synthetic preference dataset. (1) we first train the sequence-wise reward model $r(\rvx,\rvy)$ on the synthetic preference dataset, then (2) we apply the FUDGE algorithm to train the prefix scorer $V_{\theta}$ with the sequence-wise reward model $r$ as the reward function.

Given the implicit and explicit value functions, we use them to \textbf{steer the large pre-trained language models without additional training}. 
Since token-wise sampling only supports steering the model sharing the same vocabulary as the base model,
here we only study on the \text{gpt2} family models with different scales: \texttt{gpt2-mudium} (345M parameters), \texttt{gpt2-large} (774M parameters) and \texttt{gpt2-xl} (1.5B parameters).
Eventually, since we have access to the gold reward model, responses can be fairly evaluated on the test split of prompts using this gold reward model.

\begin{figure*}[ht!]
    \centering
    \includegraphics[width=\textwidth]{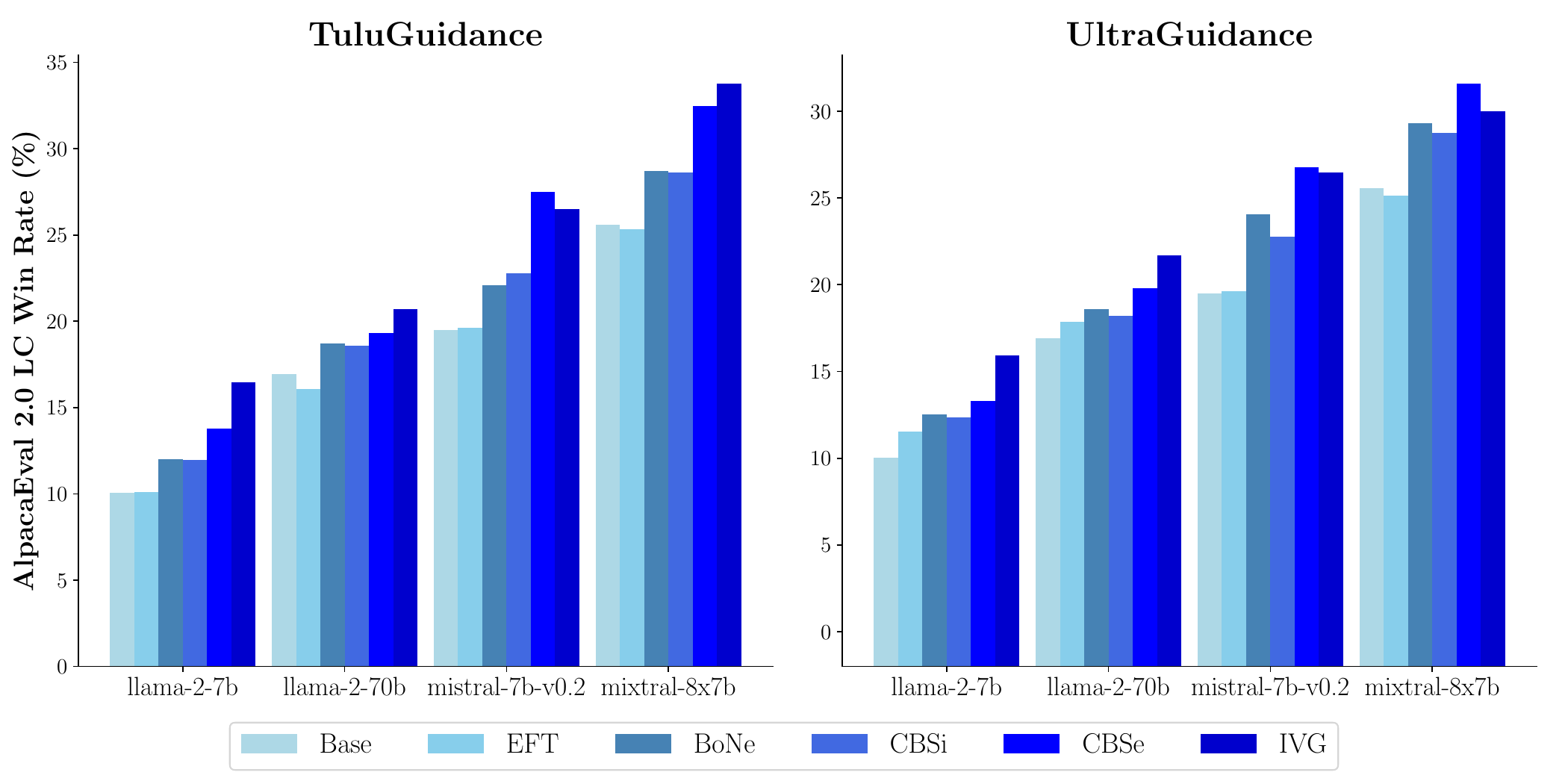}
    \caption{
        \textbf{The length-controlled win rates~(LC Win Rates)} against gpt-4-turbo for various instruction-tuned models under TuluGuidance and UltraGuidance.
        BoNe denote BoN ($N=16$) with explicit rewards, respectively;
        EFTi ($\beta^*$) denotes the results with $\beta_i=1.0$ for Tulu guidance and $\beta_i=1.5$ for Ultra guidance which are the best parameters we evaluated on \texttt{Llama-2-7b-chat-hf}.
        CBSi and CBSe denote the results with implicit and explicit value functions with $W,K,L=2,2,30$.
        IVG denotes the best results with $W,K,L=2,2,30$ with $\beta=\beta_i$.
        BoNi and EFTe are not shown due to their weak performance.
        More results are shown in appendix \ref{app:sec:extened-exp-results}.
    }
    \label{fig:performance_of_alpaca_eval}
\vspace{-5pt}
\end{figure*}

\paragraph{Results.}
Figure~\ref{fig:performance_of_imdb_summarization} demonstrates IVG's outstanding performance in both controlled-sentiment generation and summarization tasks. We find that \textbf{IVG achieves the best performance among all the baselines}, showing the effectiveness of the proposed method. 
To assess the effectiveness of IVG, we examined how different combinations of implicit and explicit value functions perform in two tasks, as depicted in Figure~\ref{fig:heatmap_of_imdb_summarization}. 
In chunk-level beam search, the explicit value function significantly enhances performance in both tasks, whereas the implicit value function shows lesser improvements. Conversely, in token-wise sampling, the implicit value function notably boosts performance in the controlled-sentiment generation task, but the explicit value function has a minimal impact. 
This distinction likely arises because the controlled-sentiment generation task primarily requires adjustments at the token level (e.g., "dislike" $\rightarrow$ "like"), whereas summarization demands a focus on broader contextual information. The results suggest that the explicit value function excels in chunk-level beam search, while the implicit value function performs better in token-wise sampling. By integrating both value functions, IVG achieves superior performance in both tasks.

\subsection{Instruction-Following}\label{subsec:exp-instruction-following}

\paragraph{Setup.} We evaluate IVG on AlpacaEval 2.0~\citep{dubois2024length}, a single-turn instruction-following benchmark comprising 805 prompts from various open-source datasets. Here, unlike steering \textit{pre-trained} models (e.g., \texttt{Llama-2-7b}), we utilize \textit{instruction-tuned} models (e.g., \texttt{Llama-2\-7b-chat}) due to their need for additional alignment as per~\citep{ji2024aligner}.

For smaller models, we adopt two strategies to derive implicit and explicit value functions: 
(1) \textbf{Tulu guidance}: Utilizing \texttt{tulu-2-dpo-7b} and its baseline \texttt{tulu-2-7b} for the implicit function, and training \texttt{FsfairX-LLaMA3-RM-v0.1} on the \texttt{UltraFeedback} dataset for the explicit function.
(2) \textbf{Ultra guidance}: Fine-tuning \texttt{Llama-2-7b} via Direct Preference Optimization (DPO)~\citep{rafailov2024direct} on \texttt{UltraFeedback} for both implicit and explicit functions.
All the models use the Llama-2 tokenizer.

Target instruction-tuned models include \texttt{Llama-2-7b-chat-hf}, \texttt{Llama-2-70b-chat-hf}, \texttt{Mistral-7B-Instruct-v0.2} and \texttt{Mixtral\-8x7B-Instruct-v0.1}.
To manage computational costs, we refrained from directly fine-tuning large models. We explored various valid combinations of implicit and explicit value functions, such as CBSe for chunk-level beam search with explicit value function. 
Language model responses are evaluated by their length-controlled win rates (LC WR) against \texttt{gpt-4-turbo}, with \texttt{gpt-4-turbo} serving as the judge.

\paragraph{Results.}
Figure~\ref{fig:performance_of_alpaca_eval} demonstrates that consistently performs well. Notably, applying an explicit value function to chunk-level beam search significantly enhances outcomes, whereas an implicit value function improves results when applied to token-wise sampling, albeit less effectively for larger models such as \texttt{Mixtral-8x7B\-Instruct-v0.1}. These findings confirm the theoretical trade-offs between explicit and implicit value functions.

\paragraph{Ablation.}

\begin{figure}[htbp]
    \centering
    \includegraphics[width=0.5\textwidth]{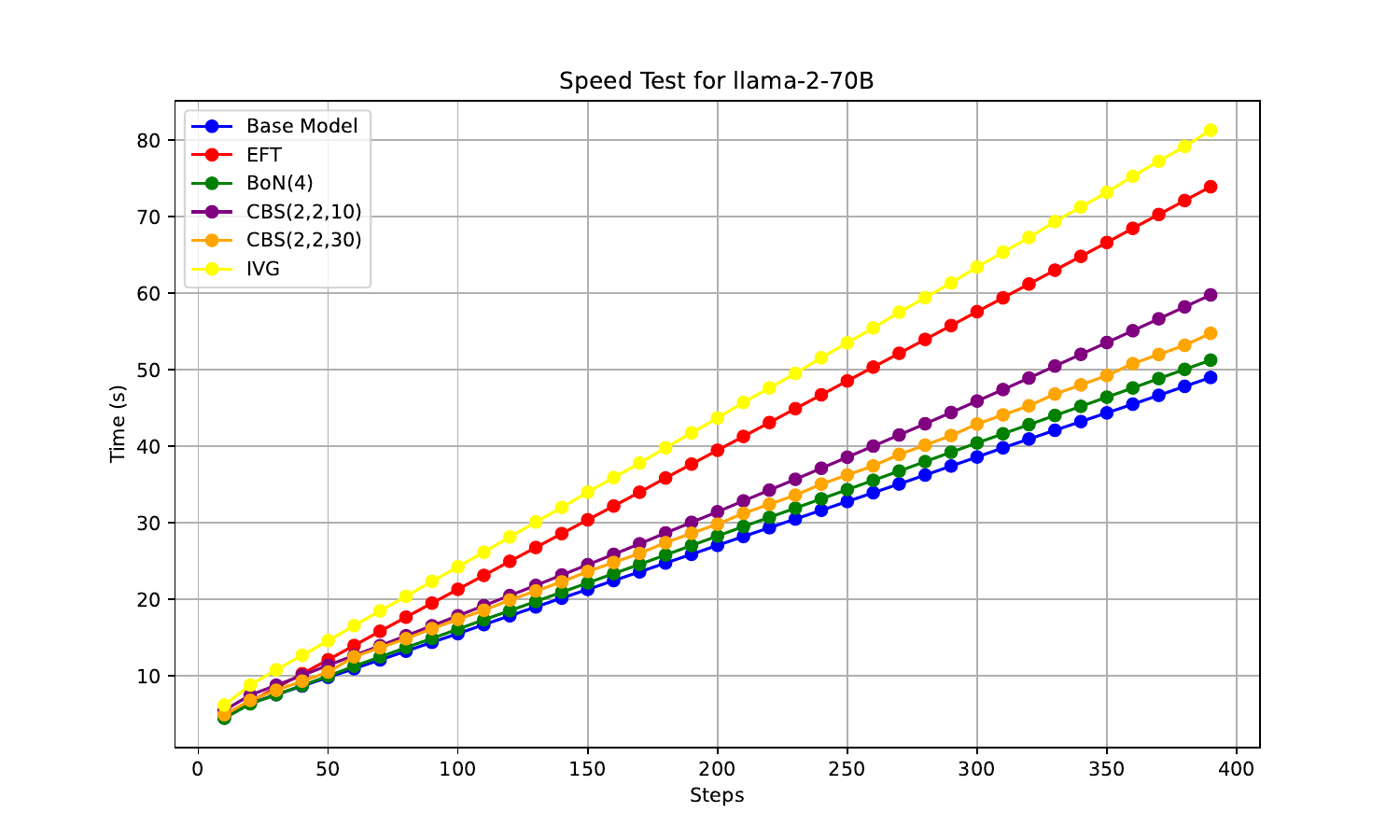}
    \caption{Inference Speed Comparison for \texttt{Llama-2\-70B-chat-hf}. The figure illustrates the inference time across various methods over a range of steps. }
    \label{fig:speed_test}
\end{figure}

We present an ablation study evaluating the inference speed of various methods applied to instruction-following tasks using a single sample.
We used TuluGuidance to guide the \texttt{Llama-2-70B-chat-hf}, and the results are illustrated in Figure \ref{fig:speed_test}.
Our empirical findings corroborate the theoretical analysis. The additional time overhead associated with IVG is primarily due to the extra inference steps required by both the implicit and explicit reward models. Notably, the implicit reward model incurs significantly higher inference costs relative to the explicit model, which can be attributed to the markedly higher forward pass frequency inherent to the implicit approach.
The Chunk-level Beam Search (CBS) with hyperparameters $W=2$, $K=2$, and $L=30$ demonstrates a substantial efficiency advantage over Emulator Fine-Tuning (EFT). CBS achieves superior optimization results while incurring only few additional time costs. In contrast, EFT, despite its higher time overhead, yields only modest improvements across certain models. Therefore, in practice, employing Chunk-Level Beam Search alone may represents a more efficient choice.

\section{Discussion}\label{sec:discussion}

We have presented Integrated Value Guidance (IVG), a method that combines implicit and explicit value functions, applied to token-wise sampling and chunk-level beam search. We conducted experiments on synthetic tasks and instruction-following task and found that IVG achieved the best performance in both tasks. We also explored the performance of different combinations of implicit and explicit value functions in the two tasks and found that the explicit value function applied to chunk-level beam search can significantly improve the results, while the implicit value function applied to token-wise sampling can improve the results. IVG combines the advantages of both, so it achieves the best performance in both tasks.

\paragraph{Limitations.}
Our work primarily focuses on enhancing the alignment capabilities of large language models through the integration of implicit and explicit value functions. This approach introduces several complex questions that extend beyond the current scope of our research:

\begin{enumerate}
    \item Our methodologies have been limited to using the DPO~\citep{rafailov2024direct} and FUDGE~\cite{mudgal2023controlled,yang-klein-2021-fudge} algorithms for training and deriving the implicit and explicit value functions. It remains unclear whether incorporating other large model alignment strategies or offline reinforcement learning algorithms for token-wise sampling and chunk-level beam search might influence our findings. This aspect warrants additional experimental investigation.
    \item Although we have detailed the empirical outcomes associated with the Implicit and Explicit Value Functions, a theoretical framework that explicates these results is conspicuously absent. Developing a theoretical understanding to underpin these empirical findings is an essential next step for further research.
\end{enumerate}

\section{Acknowledgement}\label{sec:discussion}

This work is supported in part by the National Key R\&D Program of China~(NO.2022ZD0160102).

\clearpage
\bibliography{
references/alignment,
references/search,
references/proxy_tuning,
references/models,
references/benchmarks,
references/others
}

\clearpage
\appendix
\section{Experimental Setup Details}\label{app:sec:exp-setup-details}

We adopt the experimental setup from \citet{zhou2024weak}.

\subsection{Controlled-Sentiment Generation \& Summarization}\label{app:subsec:exp-setup-details-synthetic}

\subsubsection{Model Specification}\label{app:subsubsec:synthetic-model}
The models are specified in the following table:

\begin{table}[ht]
\centering
\footnotesize 
\begin{tabular}{@{}l@{}} 
\toprule
\textbf{Models and Links} \\
\midrule
\texttt{gpt2} (124M)~\cite{radford2019language} \\
\url{https://huggingface.co/openai-community/gpt2} \\
\\
\texttt{gpt2-medium} (345M)~\cite{radford2019language} \\
\url{https://huggingface.co/openai-community/gpt2-medium} \\
\\
\texttt{gpt2-large} (774M)~\cite{radford2019language} \\
\url{https://huggingface.co/openai-community/gpt2-large} \\
\\
\texttt{gpt2-xl} (1.5B)~\cite{radford2019language} \\
\url{https://huggingface.co/openai-community/gpt2-xl} \\
\bottomrule
\end{tabular}
\caption{Models and their links}

\end{table}

\subsubsection{Hyperparameters Specification}\label{app:subsubsec:synthetic-hyperp}
In our approach, we use fixed hyperparameters across all tested models to ensure consistency. During decoding, we set the temperature to $T=0.7$, with $\text{top-k} = \text{None}$ and $\text{top-p} = 1.0$. For chunk-level beam search, the parameters are configured as follows: beam width $W = 4$, successors per state $K = 4$, and chunk length $L = 5$. To maintain computational fairness, we set the number of samples $N$ to 16 for the BoN sampling. For EFT, we report the best results obtained from $\beta \in \{0.25, 0.5, 1, 2\}$.

\subsubsection{Compute Resources}\label{app:subsubsec:synthetic-compute}
Evaluation occurs over 1000 test prompts using a single NVIDIA A100 GPU.


\subsubsection{Gold Reward Models}\label{app:subsubsec:synthetic-grm}

We follow a synthetic setup where gold reward models simulate human evaluations by generating binary preference labels~\cite{gao2023scaling, lightman2023let, rafailov2024direct}.  

For controlled-sentiment generation, we utilize the publicly accessible \href{https://huggingface.co/lvwerra/distilbert-imdb}{\texttt{distilbert-imdb}} as our gold reward model $r_{\text{gold}}$. The \href{https://huggingface.co/lvwerra/distilbert-imdb}{\texttt{distilbert-imdb}} is a fine-tuned classifier $p$ on the \href{https://huggingface.co/datasets/stanfordnlp/imdb}{\texttt{imdb}} dataset~\cite{maas-etal-2011-learning}, designed to assess the sentiment of movie reviews.
We define the gold reward $r_{\text{gold}}$ as $\log p(\text{positive} \, | \, x, y) - \log p (\text{negative} \, | \, x, y)$, promoting positive sentiment reviews.
Synthetic preferences are collected using the truncated movie reviews as prompts $\rvx$, and pairwise completions from \href{https://huggingface.co/lvwerra/gpt2-imdb}{\texttt{gpt2-imdb}}, ranked by $p(\rvy_1 \succ \rvy_2 \mid \rvx) = \sigma (r_{\text{gold}}(\rvx, \rvy_1) - r_{\text{gold}}(\rvx, \rvy_2))$.

For the summarization task, we fit a reward model on the \href{https://huggingface.co/datasets/openai/summarize_from_feedback}{\texttt{summarize\_from\_feedback}} dataset~\cite{stiennon2020learning} as our gold reward model $r_{\text{gold}}$. 
This model is specifically fine-tuned from \texttt{Llama-2-7b} with a linear projection head and binary cross-entropy loss. The training parameters include a batch size of $32$, a learning rate of \texttt{1e-5} for the projection head, and \texttt{5e-6} for the other parameters, conducted over one epoch with a cosine learning rate schedule.
Synthetic preferences are generated by relabeling pairwise responses in the original dataset, using $p(\rvy_1 \succ \rvy_2 \mid \rvx) = \sigma (r_{\text{gold}}(\rvx, \rvy_1) - r_{\text{gold}}(\rvx, \rvy_2))$.

Both gold reward models exhibit high validation accuracies of $0.928$ and $0.736$, respectively, indicating a strong alignment with human judgment.

\subsubsection{Direct Tuning Details}\label{app:subsubsec:synthetic-tuning-details}
Direct tuning on the synthetic preferences $\mathcal{D} = \{(\rvx, \rvy^w, \rvy^l)_i \}_{i=1}^N$ involves two stages: Supervised Fine-Tuning (\texttt{SFT}) and Direct Preference Optimization (\texttt{DPO})~\cite{rafailov2024direct}. During \texttt{SFT}, models are trained on both selected and rejected responses using a batch size of $64$, a learning rate of \texttt{2e-5}, and a cosine learning rate schedule over one epoch. During \texttt{DPO}, we use a $\beta=0.1$, batch size of $256$, a learning rate of \texttt{1e-6}, and a cosine learning rate schedule over one epoch.

\subsubsection{Prompt Template for Sampling from Base Models}\label{app:subsubsec:synthetic-prompt-template}
For sentiment-controlled generation, we use a zero-shot prompt:

\begin{center}

\verb|Here is a movie review from imdb: {prompt}|

\end{center}

For summarization, we use a two-shot prompt (the exemplars are selected arbitrarily):
\begin{verbatim}

{examplar[1].prompt}TL;DR: {examplar[1].response}
{examplar[2].prompt}TL;DR: {examplar[2].response}
   {prompt}TL;DR:

\end{verbatim}

\subsection{Instruction Following}\label{app:subsec:exp-setup-details-instruction-following}
\subsubsection{Model Specification}\label{app:subsubsec:instruction-following-model}
The following table lists the models and their corresponding links.

\begin{table}[ht]
\centering
\resizebox{\columnwidth}{!}{ 
\begin{tabular}{@{}l@{}}
\toprule
\textbf{Models and Links} \\
\midrule
\texttt{tulu-2-dpo-7b}~\cite{ivison2023camels} \\
\url{https://huggingface.co/allenai/tulu-2-dpo-7b} \\
\\
\texttt{tulu-2-7b}~\cite{ivison2023camels} \\
\url{https://huggingface.co/allenai/tulu-2-7b} \\
\\
\texttt{Llama-2-7b-chat}~\cite{touvron2023llama2} \\
\url{https://huggingface.co/meta-llama/Llama-2-7b-chat-hf} \\
\\
\texttt{Llama-2-70b-chat}~\cite{touvron2023llama2} \\
\url{https://huggingface.co/meta-llama/Llama-2-70b-chat-hf} \\
\\
\texttt{Mistral-7B-Instruct-v0.2}~\cite{jiang2023mistral} \\
\url{https://huggingface.co/mistralai/Mistral-7B-Instruct-v0.2} \\
\\
\texttt{Mixtral-8x7B-Instruct-v0.1}~\cite{mixtral2023} \\
\url{https://huggingface.co/mistralai/Mixtral-8x7B-Instruct-v0.1} \\
\\

\texttt{FsfairX-LLaMA3-RM-v0.1}~\cite{dong2023raft,xiong2024iterative}\\
\url{https://huggingface.co/sfairXC/FsfairX-LLaMA3-RM-v0.1}\\
\bottomrule
\end{tabular}
}
\caption{Models and their links}

\end{table}

\subsubsection{Hyperparameters Specification.}\label{app:subsubsec:instruction-following-hyperparam}
We use fixed hyperparameters across all tested models. We use temperature $T=0.7$, $\text{top-k} = 50$ and $\text{top-p} = 1.0$.
For chunk-level beam search, the parameters are configured as follows: beam width $W = 2$, successors per state $K = 2$, and chunk length $L = 30$. To maintain computational fairness, we set the number of samples $N$ to 4 for the BoN sampling. For EFT, we report the results of fixed $\beta_e,\beta_i$, which are the best parameters evaluated on \texttt{Llama-2-7b-chat-hf}.

\subsubsection{Compute Resources Specification.}\label{app:subsubsec:instruction-following-compute}
Models are evaluated on 805 test prompts. Model inference takes place on one single NVIDIA A100 GPU for 7B models and on four for others.

\section{Extended Experimental Results}\label{app:sec:extened-exp-results}

Table~\ref{tab:tulu-guidance} and Table~\ref{tab:ultra-guidance} present complete experimental results under the instruction following task. EFT~\cite{mitchell2023emulator} represents applying token-wise sampling during generation and CBS~\cite{zhou2024weak} represents applying chunk-level beam search during generation. The suffix "i" and "e" indicates that implicit and explicit value functions. Note that the proposed Integrated Value Guidance (IVG) is shown as EFTi $(\beta_i)$,CBSe.

In addition to \texttt{gpt-4-turbo} evaluations, we evaluate response by using two top-rank reward models from RewardBench~\cite{lambert2024rewardbench}: \href{https://huggingface.co/openbmb/UltraRM-13b}{\texttt{UltraRM-13b}}~\citep{cui2023ultrafeedback} and \href{https://huggingface.co/Nexusflow/Starling-RM-34B}{\texttt{Starling-RM-34B}}~\citep{starling2023}. SRM denotes the scores evaluated by \texttt{Starling-RM-34B} and URM denotes the scores evaluated by \texttt{UltraRM-13b}.

\renewcommand{\arraystretch}{1.2}
\begin{table*}[htbp]
\centering
\resizebox{1.3\columnwidth}{!}{%
\begin{tabular}{lrrrr}
\toprule
\textbf{Models} & \textbf{SRM $(\uparrow)$} & \textbf{URM $(\uparrow)$} & \textbf{LC WR (\%)}  & \textbf{WR (\%)} \\
\midrule
\rowcolor{blue!10}
\multicolumn{5}{c}{Tulu Guidance}    \\
\midrule
\multicolumn{5}{c}{\texttt{Llama-2-7b-chat}}  \\
Base & $-5.83$ & $1.24$ & $10.04$ & $10.16$ \\
EFTe $(\beta_e)$ & $-5.82$ & $1.26$ & N/A & N/A \\
EFTi $(\beta_i)$ & $-5.68$ & $1.53$ & $10.09$ & $10.79$ \\
BoNe  & $-5.53$ & $2.04$ & $12.00$ & $12.82$ \\
CBSi  & $-5.61$ & $1.78$ & $11.96$ & $13.06$ \\
EFTi $(\beta_i)$,CBSi  & $-5.56$ & $1.86$ & $11.76$ & $12.82$ \\
CBSe  & $-5.30$ & $2.58$ & $13.76$ & $15.01$ \\
EFTi $(\beta_i)$,CBSe  & \textbf{-5.18} & \textbf{2.95} & \textbf{16.46} & \textbf{18.00} \\
\midrule
\multicolumn{5}{c}{\texttt{Llama-2-70b-chat}} \\
Base & $-5.61$ & $1.94$ & $16.93$ & $15.71$ \\
EFTe $(\beta_e)$ & $-5.57$ & $1.87$ & N/A & N/A \\
EFTi $(\beta_i)$ & $-5.47$ & $2.16$ & $16.07$ & $15.44$ \\
BoNe  & $-5.30$ & $2.58$ & $18.70$ & $18.38$ \\
CBSi  & $-5.43$ & $2.29$ & $18.57$ & $17.97$ \\
EFTi $(\beta_i)$,CBSi  & $-5.33$ & $2.50$ & $17.38$ & $17.40$ \\
CBSe  & $-5.26$ & $2.79$ & $19.32$ & $19.06$ \\
EFTi $(\beta_i)$,CBSe  & \textbf{-5.11} & \textbf{3.08} & \textbf{20.72} & \textbf{21.26} \\
\midrule
\multicolumn{5}{c}{\texttt{Mistral-7B-Instruct-v0.2}} \\
Base & $-5.72$ & $2.05$ & $19.51$ & $16.27$ \\
EFTe $(\beta_e)$ & $-11.12$ & $-8.01$ & N/A & N/A \\
EFTi $(\beta_i)$ & $-5.64$ & $2.27$ & $19.62$ & $16.52$ \\
BoNe  & $-5.42$ & $2.89$ & $22.10$ & $18.79$ \\
CBSi  & $-5.47$ & $2.74$ & $22.77$ & $19.34$ \\
EFTi $(\beta_i)$,CBSi  & $-5.43$ & $2.77$ & $20.47$ & $19.02$ \\
CBSe  & $-5.19$ & $3.54$ & \textbf{27.50} & $24.21$ \\
EFTi $(\beta_i)$,CBSe  & \textbf{-5.05} & \textbf{3.70} & $26.51$ & \textbf{25.15} \\
\midrule
\multicolumn{5}{c}{\texttt{Mixtral-8x7B-Instruct-v0.1}} \\
Base & $-5.67$ & $1.89$ & $25.58$ & $19.56$ \\
EFTe $(\beta_e)$ & $-8.85$ & $-3.58$ & N/A & N/A \\
EFTi $(\beta_i)$ & $-5.55$ & $2.05$ & $25.32$ & $20.25$ \\
BoNe  & $-5.33$ & $2.68$ & $28.71$ & $23.61$ \\
CBSi  & $-5.43$ & $2.38$ & $28.62$ & $22.85$ \\
EFTi $(\beta_i)$,CBSi  & $-5.33$ & $2.56$ & $27.67$ & $22.50$ \\
CBSe  & $-5.14$ & $3.14$ & $32.49$ & $27.69$ \\
EFTi $(\beta_e)$,CBSe  & \textbf{-5.12} & \textbf{3.21} & \textbf{33.75} & \textbf{28.30} \\
\midrule
\end{tabular}
}
\vspace{.5cm}
\caption{
\textbf{Instruction following performance under the Tulu guidance.} $\beta_i,\beta_e= 1.0,1.0$. All CBS shows the results with $W,K,L=2,2,30$. BoN shows the results with $N=4$. }
\label{tab:tulu-guidance}
\end{table*}

\renewcommand{\arraystretch}{1.22}
\begin{table*}[htbp]
\centering
\resizebox{1.3\columnwidth}{!}{%
\begin{tabular}{lrrrr}
\toprule
\textbf{Models} & \textbf{SRM $(\uparrow)$} & \textbf{URM $(\uparrow)$} & \textbf{LC WR (\%)}  & \textbf{WR (\%)} \\
\midrule
\rowcolor{blue!10}
\multicolumn{5}{c}{Ultra Guidance} \\
\midrule
\multicolumn{5}{c}{\texttt{Llama-2-7b-chat}} \\
Base & $-5.83$ & $1.24$ & $10.04$ & $10.16$ \\
EFTe $(\beta_e)$ & $-5.81$ & $1.30$ & N/A & N/A \\
EFTi $(\beta_i)$ & $-5.64$ & $1.68$ & $11.53$ & $12.04$ \\
BoNe  & $-5.61$ & $1.83$ & $12.51$ & $12.78$ \\
CBSi  & $-5.55$ & $1.81$ & $12.36$ & $13.04$ \\
EFTi $(\beta_i)$,CBSi  & $-5.42$ & $2.23$ & $13.58$ & $14.49$ \\
CBSe  & $-5.47$ & $2.19$ & $13.31$ & $13.81$ \\
EFTi $(\beta_i)$,CBSe  & \textbf{-5.29} & \textbf{2.71} & \textbf{15.92} & \textbf{16.70} \\
\midrule
\multicolumn{5}{c}{\texttt{Llama-2-70b-chat}} \\
Base & $-5.61$ & $1.94$ & $16.93$ & $15.71$ \\
EFTe $(\beta_e)$ & $-5.62$ & $1.89$ & N/A & N/A \\
EFTi $(\beta_i)$ & $-5.49$ & $2.16$ & $17.85$ & $17.07$ \\
BoNe  & $-5.39$ & $2.40$ & $18.58$ & $17.68$ \\
CBSi  & $-5.35$ & $2.47$ & $18.20$ & $17.70$ \\
EFTi $(\beta_i)$,CBSi  & $-5.29$ & $2.60$ & $19.21$ & $18.70$ \\
CBSe  & $-5.36$ & $2.57$ & $19.82$ & $18.54$ \\
EFTi $(\beta_i)$,CBSe  & \textbf{-5.25} & \textbf{2.84} & \textbf{21.69} & \textbf{21.06} \\
\midrule
\multicolumn{5}{c}{\texttt{Mistral-7B-Instruct-v0.2}} \\
Base & $-5.72$ & $2.05$ & $19.51$ & $16.27$ \\
EFTe $(\beta_e)$ & $-7.36$ & $-0.85$ & N/A & N/A \\
EFTi $(\beta_i)$ & $-5.62$ & $2.25$ & $19.64$ & $17.70$ \\
BoNe  & $-5.55$ & $2.64$ & $24.04$ & $19.18$ \\
CBSi  & $-5.44$ & $2.81$ & $22.78$ & $19.70$ \\
EFTi $(\beta_i)$,CBSi  & $-5.36$ & $2.88$ & $21.71$ & $20.49$ \\
CBSe  & $-5.38$ & $3.12$ & \textbf{26.78} & $22.23$ \\
EFTi $(\beta_i)$,CBSe  & \textbf{-5.26} & \textbf{3.36} & $26.46$ & \textbf{23.60} \\
\midrule
\multicolumn{5}{c}{\texttt{Mixtral-8x7B-Instruct-v0.1}} \\
Base & $-5.67$ & $1.89$ & $25.58$ & $19.56$ \\
EFTe $(\beta_e)$ & $-5.78$ & $1.74$ & N/A & N/A \\
EFTi $(\beta_i)$ & $-5.59$ & $2.02$ & $25.11$ & $20.25$ \\
BoNe  & $-5.43$ & $2.46$ & $29.29$ & $23.35$ \\
CBSi  & $-5.40$ & $2.54$ & $28.76$ & $23.09$ \\
EFTi $(\beta_i)$,CBSi  & $-5.33$ & $2.59$ & $29.92$ & \textbf{25.27} \\
CBSe  & $-5.33$ & $2.78$ & \textbf{31.57} & $25.18$ \\
EFTi $(\beta_i)$,CBSe  & \textbf{-5.30} & \textbf{2.78} & $29.99$ & $24.24$ \\
\midrule
\end{tabular}
}
\vspace{.5cm}
\caption{
\textbf{Instruction following performance under the Ultra guidance.} $\beta_i,\beta_e= 1.5,1.0$. All CBS shows the results with $W,K,L=2,2,30$. BoN shows the results with $N=4$.}
\label{tab:ultra-guidance}
\end{table*}

\end{document}